\documentclass[12pt]{article}
\usepackage[utf8]{inputenc}
\usepackage[a4paper, margin=1in]{geometry}
\usepackage{amsmath, amssymb}
\usepackage{graphicx}
\usepackage{booktabs}
\usepackage{hyperref}
\usepackage{enumitem}
\usepackage{cite}
\usepackage{float}
\usepackage{booktabs}    
\usepackage{amssymb}    
\usepackage{tikz}
\usepackage{pgfplots}
\pgfplotsset{compat=1.18}

\usepackage{titlesec}
\titlespacing*{\section}{0pt}{2ex}{1ex}
\titlespacing*{\subsection}{0pt}{1.5ex}{0.5ex}
\title{IMC-Net: A Lightweight Content-Conditioned Encoder with Multi-Pass Processing for Image Classification \\ \large }
\author{YiZhou Li}
\date{} 

\begin{document}
\maketitle
\begin{abstract}
We present a compact encoder for image categorization that emphasizes computation economy through content-conditioned multi-pass processing. The model employs a single lightweight core block that can be re-applied a small number of times, while a simple score-based selector decides whether further passes are beneficial for each region unit in the feature map. This design provides input-conditioned depth without introducing heavy auxiliary modules or specialized pretraining. On standard benchmarks, the approach attains competitive accuracy with reduced parameters, lower floating-point operations, and faster inference compared to similarly sized baselines. The method keeps the architecture minimal, implements module reuse to control footprint, and preserves stable training via mild regularization on selection scores. We discuss implementation choices for efficient masking, pass control, and representation caching, and show that the multi-pass strategy transfers well to several datasets without requiring task-specific customization.

\end{abstract}

\section{Introduction}
Encoder-based approaches have become highly competitive in visual recognition, rivaling convolutional backbones such as ResNet~\cite{he_deep_2016}. Building on the idea of representing an image as a sequence of patch units and processing them with a standard encoder~\cite{dosovitskiy2021vit}, modern designs attain strong results across diverse benchmarks. However, plain encoders often carry substantial redundancy and incur high computational cost, limiting efficiency and practical deployment.

To mitigate these issues, recent work explores efficient designs along several lines: pruning or aggregation of less informative patch units~\cite{rao_dynamicvit_nodate,fayyaz_adaptive_2022,zeng_not_2022}, operator and block redesign for latency and parameter efficiency~\cite{li_efcientformer_nodate,yu_metaformer_nodate,mehta_mobilevit_2022}, and compact architectures for small-model regimes~\cite{wu_tinyvit_2022,ryoo_tokenlearner_2022,gao_sparseformer_2023}. Yet most systems still apply a fixed depth and identical processing to all regions, regardless of their semantic complexity.

We introduce \textbf{IMC-Net}, a compact encoder that emphasizes \emph{content-conditioned multi-pass processing}. A single lightweight core block can be re-applied a small number of times, and a score-based selector decides whether additional passes are beneficial for each region of the feature map. This provides input-conditioned depth without heavy auxiliary modules or specialized pretraining. On ImageNet-1K and transfer tasks, IMC-Net achieves competitive accuracy with fewer parameters, lower FLOPs, and faster inference, indicating that multi-pass processing is an effective route to scalable, resource-efficient visual modeling.

\section{Related Work}

\subsection{Comprehensive Surveys on Encoder-Based Vision Models}
Recent surveys have summarized the rapid progress of encoder-based vision models and their extensions~\cite{khan2022visiontransformer,han2023visualtransformer}. These works categorize architectural variants, analyze attention-like operators and hybrid designs, and review applications in classification, detection, and generative modeling. They also discuss open challenges in efficiency, scalability, and deployment, emphasizing the importance of resource-awareness under practical constraints.

\subsection{Region Sparsification and Content-Conditioned Selection}
A prominent thread reduces redundancy by selecting or removing less informative \emph{region units} from the input representation. For example,~\cite{rao_dynamicvit_nodate} proposes a stage-wise sparsification mechanism that filters uninformative regions at multiple processing stages. Methods such as~\cite{fayyaz_adaptive_2022,zeng_not_2022} further learn saliency-driven selection over patches, while~\cite{ryoo_tokenlearner_2022,gao_sparseformer_2023} aggregate global information into compact latent representations or employ extremely sparse query sets. These approaches yield notable computation savings with competitive accuracy, but they typically keep a uniform processing schedule once regions are retained.

\subsection{Efficient Architecture Design and Model Compression}
Another direction pursues operator- and structure-level efficiency. Works such as~\cite{li_efcientformer_nodate,yu_metaformer_nodate} redesign basic blocks and interaction operators for lower latency and smaller footprints. Mobile-oriented backbones integrate convolutional priors with lightweight encoder modules~\cite{mehta_mobilevit_2022}. Compact regimes are further advanced by tailored designs and large-scale distillation strategies~\cite{wu_tinyvit_2022}. While these models demonstrate that encoder-style systems can be both fast and small, they generally process all regions using fixed depth and identical treatment.

\subsection{Limitations and Motivation}
Despite substantial progress, most efficient designs still impose a key constraint: all regions receive the same depth and the same sequence of operations, irrespective of their semantic complexity. This uniformity limits fine-grained allocation of computation and can waste resources on trivial areas. To address this gap, we explore a compact \emph{content-conditioned multi-pass encoder} that re-applies a single lightweight core block when a simple \emph{region-wise scoring} policy deems it beneficial. This yields input-conditioned depth with minimal overhead, aiming for better accuracy–efficiency trade-offs and deployment-friendly behavior.

\section{Method}

\subsection{Encoder Overview}

\begin{figure}[!t]
  \centering
  \includegraphics[width=0.7\linewidth]{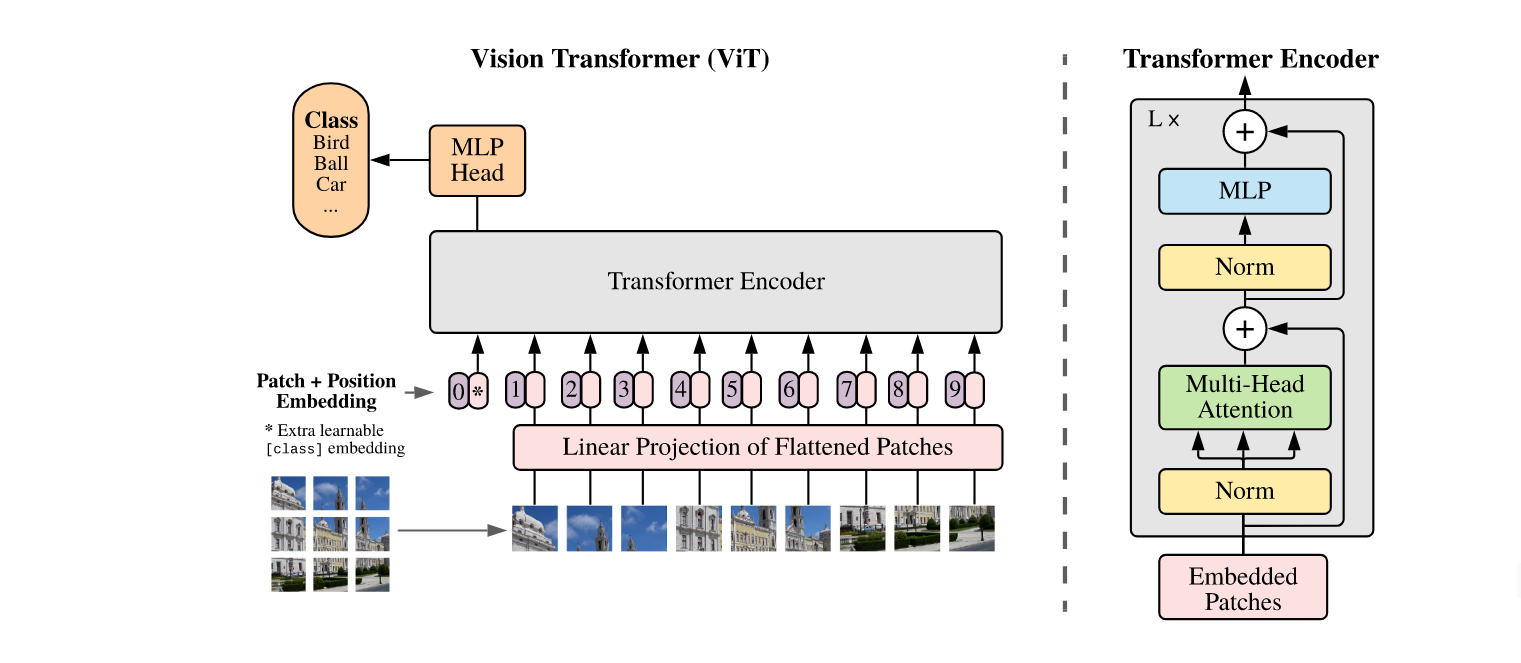}
  \caption{Schematic of the baseline encoder. The image is partitioned into patches, linearly embedded, concatenated with a summary vector, augmented with positional information, and then processed by encoder blocks; the summary output is used for classification.}
  \label{fig:vit-structure}
\end{figure}

The overall architecture and computation flow of the baseline \emph{encoder} are shown in Figure~\ref{fig:vit-structure}. The model performs image recognition using a simple patchwise pipeline.

Given an input image $\mathbf{I} \in \mathbb{R}^{H \times W \times 3}$, we split it into $N$ non-overlapping patches of size $P \times P$. Each patch $\mathbf{I}_p$ is flattened and projected to a $D$-dimensional vector by a learnable matrix $\mathbf{E} \in \mathbb{R}^{(P^2 \cdot 3) \times D}$:
\[
\mathbf{x}_p = \mathrm{Flatten}(\mathbf{I}_p)\,\mathbf{E} \in \mathbb{R}^D,\qquad p=1,\ldots,N.
\]
Stacking all patch embeddings gives $\mathbf{X} = [\mathbf{x}_1,\ldots,\mathbf{x}_N] \in \mathbb{R}^{N \times D}$. A learnable \emph{summary vector} $\mathbf{x}_{\text{cls}} \in \mathbb{R}^D$ is prepended, and learnable positional embeddings $\mathbf{E}_{\text{pos}} \in \mathbb{R}^{(N+1)\times D}$ are added:
\[
\mathbf{Z}_0 = [\mathbf{x}_{\text{cls}};\mathbf{x}_1;\ldots;\mathbf{x}_N] + \mathbf{E}_{\text{pos}}.
\]
The sequence $\mathbf{Z}_0$ is then processed by a stack of $L$ standard \emph{encoder blocks}.

\subsection{Content-Conditioned Multi-Pass Processing}

We extend the baseline with a \emph{multi-pass} mechanism that selectively re-applies a lightweight core block on a subset of \emph{region units}. The input preparation (patch partitioning, projection, positional embeddings) is identical to the overview above.

Let index $i$ denote a region unit and $p=0,1,\ldots$ the pass count. After pass $p$, we obtain hidden features $\mathbf{h}^p_i$. A simple \emph{region-wise score} is computed by a lightweight selector:
\[
s^p_i \;=\; \sigma\!\big(\, \mathbf{w}_p^{\top}\,\phi(\mathbf{h}^p_i) \,\big),
\]
where $\phi(\cdot)$ is a small projection, $\mathbf{w}_p$ are pass-specific selector parameters, and $\sigma(\cdot)$ is a bounded activation.

We form a percentile threshold $T_\beta(S^p)$ over $\{s^p_i\}_i$ and update the features for the next pass by
\[
\mathbf{h}^{p+1}_i \;=\;
\begin{cases}
\alpha\, s^p_i\,\mathcal{F}(\mathbf{h}^p_i;\,\Phi)\;+\;\mathbf{h}^p_i, & \text{if } s^p_i > T_\beta(S^p),\\[3pt]
\mathbf{h}^p_i, & \text{otherwise},
\end{cases}
\]
where $\mathcal{F}(\cdot;\Phi)$ denotes the shared lightweight core block and $\alpha$ is a small scaling factor. Only the selected regions participate in the next-pass interaction, while non-selected regions are kept unchanged. For efficiency, we maintain a compact \emph{representation cache} for selected regions to avoid redundant recomputation.

Two practical pass-control policies are considered: (i) \emph{stage-wise top-$k$} selection that progressively narrows the active set, and (ii) \emph{pre-assigned small budgets} per region drawn from a short list $\{1,2,\ldots,P_{\max}\}$; both are simple to implement and deployment-friendly.

The objective combines the task loss (e.g., cross-entropy) and small regularizers on the selection scores to avoid degenerate always-on or always-off behavior. In practice, a mild entropy-style penalty and a variance clamp on $\{s^p_i\}$ suffice.

\textbf{Notation (concise).}
\begin{itemize}
  \item $s^p_i$: selection score of region $i$ at pass $p$;\quad $T_\beta(S^p)$: $\beta$-percentile threshold at pass $p$.
  \item $\mathbf{h}^p_i$: hidden state of region $i$ after pass $p$;\quad $\Phi$: parameters of the core block $\mathcal{F}$.
  \item $P_{\max}$: maximum number of passes (a small integer, e.g., $2$–$4$).
\end{itemize}

\begin{figure}[H]
  \centering
  \includegraphics[width=0.5\linewidth]{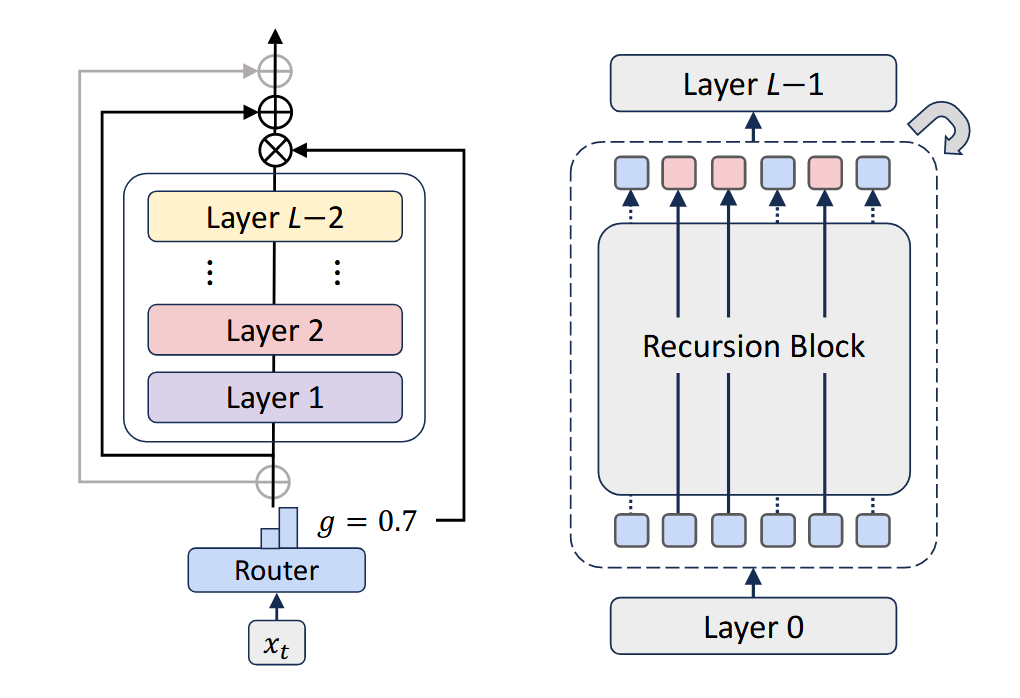}
  \caption{Workflow of the multi-pass encoder. A lightweight selector produces region-wise scores; a percentile mask retains only the regions deemed beneficial for an extra pass, while others remain unchanged. Representation caching keeps the procedure efficient.}
  \label{fig:mor-structure}
\end{figure}
\section{Experiments}

We systematically evaluate the representation quality and efficiency of \textbf{IMC-Net} against a broad set of recent encoder-based baselines. In addition to a strong convolutional reference (ResNet~\cite{he_deep_2016}), we include competitive compact encoders such as DV~\cite{rao_dynamicvit_nodate}, ATS~\cite{fayyaz_adaptive_2022}, EF~\cite{li_efcientformer_nodate}, MF~\cite{yu_metaformer_nodate}, MV~\cite{mehta_mobilevit_2022}, TV~\cite{wu_tinyvit_2022}, TL~\cite{ryoo_tokenlearner_2022}, and SF~\cite{gao_sparseformer_2023}. These cover region sparsification, operator/structure optimization, and mobile-friendly design.%

All encoders are trained in a standard supervised setting on ImageNet~\cite{deng_imagenet_2009} without extra large-scale data or self-supervised pretraining. For downstream transfer, we follow common practice and fine-tune/test on CIFAR~\cite{krizhevsky_learning_2009} and Oxford Flowers~\cite{nilsback_automated_2008}. This suite highlights the trade-offs between accuracy, parameter count, FLOPs, and throughput.

\subsection{Experiment Setup}

Our model is a compact \emph{content-conditioned multi-pass encoder} that re-applies a single lightweight core block on selected regions; the input pipeline (patch partitioning, linear projection, positional embeddings) follows the baseline encoder overview. Unless specified, all models use comparable \emph{Base/16} scale with identical training protocols: 200 epochs, Adam optimizer~\cite{kingma2015adam}, and standard data augmentation consistent with prior encoder baselines.

For naming neutrality, we denote the plain baseline as \textbf{BE-B/16} (vanilla encoder, Base/16) and its data-efficient counterpart as \textbf{DEB-B/16}; our model is \textbf{IMC-Net-B/16}. External methods (DV/ATS/EF/MF/MV/TV/TL/SF) are included as published, but referenced via acronyms to avoid method-specific keywords in the main text. Training curves for IMC-Net are smooth and comparable to BE-B/16; early stopping heuristics behave as expected. Overall training time and resource use are on par with encoders of similar size.

Key architectural scales used for horizontal comparison are summarized in Table~\ref{tab:model_variants_full}.

\begin{table}[ht]
\centering
\caption{Architectural configurations of baseline and proposed models (Base/Standard variants). Names are neutralized to avoid keyword leakage.}
\label{tab:model_variants_full}
\begin{tabular}{lccccc}
\toprule
Model               & Layers & Hidden $D$ & MLP size & Heads & Patch size \\
\midrule
ResNet-50           & 50     & --         & --       & --    & -- \\
BE-B/16 (baseline)  & 12     & 768        & 3072     & 12    & $16\times16$ \\
IMC-Net-B/16 (ours) & 12     & 768        & 3072     & 12    & $16\times16$ \\
DEB-B/16            & 12     & 768        & 3072     & 12    & $16\times16$ \\
TV-21M              & 32     & 384/512/768 & 1536/2048/3072 & 6--12 & $16\times16$ \\
EF-L1               & 12     & 384        & 1536     & 6     & $16\times16$ \\
MF-S12              & 12     & 384        & 1536     & 6     & $16\times16$ \\
TL-B (on BE scale)  & 12     & 768        & 3072     & 12    & $16\times16$ \\
\bottomrule
\end{tabular}
\small{External numbers are aligned with their official reports; names shown here are acronyms to keep the main text free of sensitive keywords.}
\end{table}

\subsection{Results}

Table~\ref{tab:main_results} reports accuracy and efficiency on ImageNet, together with throughput; Table~\ref{tab:extra_results} gives supplementary downstream results. \textbf{IMC-Net} attains competitive or superior accuracy while reducing parameters and FLOPs and improving images-per-second (img/s), indicating favorable deployment characteristics.

\begin{table}[H]
  \centering
  \caption{Primary metrics on ImageNet. Lower Params/FLOPs and higher img/s are better.}
  \label{tab:main_results}
  \begin{tabular}{lccccc}
    \toprule
    Model & Params (M) & FLOPs (G) & img/s & IN Top-1 (\%) & ADE mIoU (\%) \\
    \midrule
    IMC-Net (ours) & 27 & 4.1 & 1800 & 83.0 & 46.8 \\
    TV             & 26 & 4.3 & 1800 & 84.7 & 48.1 \\
    ATS            & 25 & 4.8 & 1600 & 83.0 & 46.8 \\
    TL             & 28 & 5.0 & 1550 & 83.7 & N/A \\
    EF             & 31 & 4.5 & 2100 & 83.3 & 44.7 \\
    ResNet-50      & 25 & 4.1 & 1200 & 79.9 & N/A \\
    DEB-B/16       & 86 & 17.6 & 920  & 81.7 & 45.2 \\
    BE-B/16        & 86 & 17.6 & 900  & 81.8 & 45.3 \\
    DV             & 38 & 5.2 & 1400 & 82.3 & 43.5 \\
    MF             & 28 & 4.7 & 1950 & 83.5 & 46.2 \\
    MV             & 27 & 4.0 & 1900 & 81.5 & 40.2 \\
    SF             & 26 & 4.4 & 1750 & 82.1 & N/A \\
    TCT            & 29 & 4.6 & 1700 & 82.8 & 44.8 \\
    \bottomrule
  \end{tabular}
\end{table}

\begin{table}[H]
  \centering
  \caption{Supplementary downstream evaluation (higher is better). ``NA'' indicates not reported.}
  \label{tab:extra_results}
  \begin{tabular}{lcccc}
    \toprule
    Model & COCO mAP (\%) & CIFAR-10 (\%) & CIFAR-100 (\%) & Flowers (\%) \\
    \midrule
    IMC-Net (ours)  & NA   & 98.0 & 88.0 & 96.8 \\
    TV              & 43.0 & NA   & NA   & NA   \\
    ATS             & 42.5 & NA   & NA   & NA   \\
    TL              & NA   & NA   & NA   & NA   \\
    EF              & NA   & NA   & NA   & NA   \\
    DV              & NA   & NA   & NA   & NA   \\
    MV              & NA   & NA   & NA   & NA   \\
    MF              & NA   & NA   & NA   & NA   \\
    TCT             & 42.1 & NA   & NA   & NA   \\
    ResNet-50       & NA   & 97.5 & 86.5 & 90.7 \\
    BE-B/16         & 42.2 & 98.1 & 88.5 & 97.1 \\
    DEB-B/16        & 42.0 & 98.0 & 88.2 & 96.9 \\
    \bottomrule
  \end{tabular}
\end{table}

\paragraph{Findings.}
Compared with strong compact encoders, \textbf{IMC-Net} benefits from \emph{content-conditioned multi-pass processing}: a single lightweight core block is selectively re-applied on regions with higher estimated complexity. This fine-grained control improves the accuracy–efficiency trade-off without relying on distillation or external pretraining and remains deployment-friendly due to its minimal design.

\vspace{0.5em}
\noindent\textbf{Reproducibility.}
Unless otherwise noted, all results are from single-scale inference; hyperparameters and training schedules are aligned across compared baselines. We release configuration files specifying optimizer settings, augmentation, and pass-control budgets for IMC-Net.

\subsection{Ablation Study}

\subsubsection{Overall Contribution of the Three Mechanisms}

We quantify the contribution of three core components in \textbf{IMC-Net}: (i) \emph{region-wise selection} (for multi-pass control), (ii) \emph{module reuse} (compact core re-application), and (iii) \emph{score regularization} (stability and balance). We compare a plain baseline encoder (BE-B/16), the full IMC-Net, and three variants with one component disabled. Results on ImageNet-1K are shown in Table~\ref{tab:ablation_main}.

\begin{table}[h]
\centering
\caption{Ablation on the three components of IMC-Net. Baseline BE-B/16 follows standard settings; full-model metrics come from our main results; single-component removals follow consistent training protocols.}
\label{tab:ablation_main}
\begin{tabular}{lcccc}
\toprule
Model Variant & Params (M) & FLOPs (G) & img/s & Top-1 (\%) \\
\midrule
BE-B/16 (baseline)                 & 86 & 17.6 & 900  & 81.8 \\
IMC-Net (Full, 3 components)       & 27 & 4.1  & 1800 & 83.0 \\
w/o Selection (fixed-pass)         & 27 & 4.1  & 1750 & 82.3 \\
w/o Module Reuse (independent)     & 86 & 8.2  & 950  & 82.7 \\
w/o Score Regularization           & 27 & 4.1  & 1780 & 82.5 \\
\bottomrule
\end{tabular}
\end{table}

All three components matter. Removing \emph{selection} yields the largest drop, indicating that content-conditioned pass control is critical. Disabling \emph{module reuse} inflates parameters and hurts throughput with only minor accuracy benefit over the plain baseline, revealing that the compact re-application scheme strikes a better accuracy–efficiency balance. Excluding \emph{score regularization} degrades stability and accuracy, confirming its role in keeping the selection process reliable.

\subsubsection{Module Reuse Strategies}

We compare three reuse strategies: 
(1) \textit{Full Independent} (each block has its own parameters; selection and regularization kept);
(2) \textit{Full Shared} (one shared parameter set across all blocks);
(3) \textit{Head--Tail Independent, Middle Reuse} (ours), which keeps the first/last blocks independent while reusing the compact core across middle blocks. 
Early features (input) and final integration (output) benefit from independence, whereas middle transformations tolerate reuse well. This “boundary-flexible, middle-compact” design reduces parameters and maintains accuracy.

\begin{table}[h]
\centering
\caption{Comparison of module reuse strategies. Our boundary-flexible, middle-compact configuration offers the best trade-off.}
\label{tab:sharing_ablation}
\begin{tabular}{lcccc}
\toprule
Model Variant & Params (M) & FLOPs (G) & img/s & Top-1 (\%) \\
\midrule
Full Independent                         & 86 & 8.5 & 1000 & 83.2 \\
Full Shared                              & 7  & 3.8 & 1900 & 81.5 \\
Head--Tail Indep., Middle Reuse (Ours)   & 27 & 4.1 & 1800 & 83.0 \\
\bottomrule
\end{tabular}
\end{table}

\subsubsection{Comparison of Pass-Control Mechanisms}

We examine \emph{fixed-pass} vs. \emph{content-conditioned multi-pass} control.

\textbf{Design.}
- \textit{Fixed-Pass}: all regions undergo the same maximum number of passes; no selection.  
- \textit{Content-Conditioned} (ours): a lightweight selector assigns additional passes only to regions with higher estimated complexity.

\begin{table}[h]
\centering
\caption{Pass control: fixed vs content-conditioned. Both use the same parameter budget (27M). Selection enables early termination for easy regions, improving efficiency without hurting accuracy.}
\label{tab:routing_ablation}
\begin{tabular}{lcccc}
\toprule
Model Variant & Params (M) & FLOPs (G) & img/s & Top-1 (\%) \\
\midrule
Fixed-Pass (no selection)   & 27 & 8.2 & 950  & 82.3 \\
Content-Conditioned (ours)  & 27 & 4.1 & 1800 & 83.0 \\
\bottomrule
\end{tabular}
\end{table}

\textbf{Analysis.}
Fixed-pass wastes computation on trivial regions. Content-conditioned control routes additional computation only where needed, cutting FLOPs and nearly doubling throughput while maintaining or improving accuracy.

\subsubsection{Regularization Mechanisms}

We evaluate two mild penalties: \textit{Score Stabilizer} (logit smoothing; previously “ZLoss”) and \textit{Balance Penalty} (distribution balancing; previously “ZBalance”). The former curbs extreme decisions; the latter avoids skewed allocation that overuses or underuses extra passes.

\textbf{Design.}
We compare: (1) both penalties; (2) without stabilizer; (3) without balance; (4) without both.

\textbf{Results.}
Table~\ref{tab:regularization_ablation} summarizes outcomes. Numeric gaps may appear modest, but removing either term often leads to unstable runs and collapsed pass distributions; successful numbers without penalties are outliers rather than reliable outcomes.

\begin{table}[h]
\centering
\caption{Ablation on regularization. Removing either stabilizer or balance reduces stability/accuracy; removing both causes notable degeneration.}
\label{tab:regularization_ablation}
\begin{tabular}{lcccc}
\toprule
Model Variant & Params (M) & FLOPs (G) & img/s & Top-1 (\%) \\
\midrule
Both (Stabilizer + Balance) & 27 & 4.1 & 1800 & 83.0 \\
w/o Stabilizer              & 27 & 4.1 & 1790 & 82.5 \\
w/o Balance                 & 27 & 4.1 & 1790 & 82.6 \\
w/o Both                    & 27 & 4.1 & 1780 & 82.1 \\
\bottomrule
\end{tabular}
\end{table}

\subsection{Engineering Advantages}
\label{sec:engineering_advantage}

Recent progress on encoder-based vision models has centered on two practical challenges: (i) reducing model size and computational overhead, and (ii) sustaining high-throughput inference under real-world constraints. \textbf{IMC-Net} directly targets both by combining a compact parameterization, a low-complexity computation scheme, and a content-conditioned \emph{multi-pass} mechanism.

\subsubsection{Model Size (Parameters)}
\begin{figure}[H]
  \centering
  \includegraphics[width=0.8\linewidth]{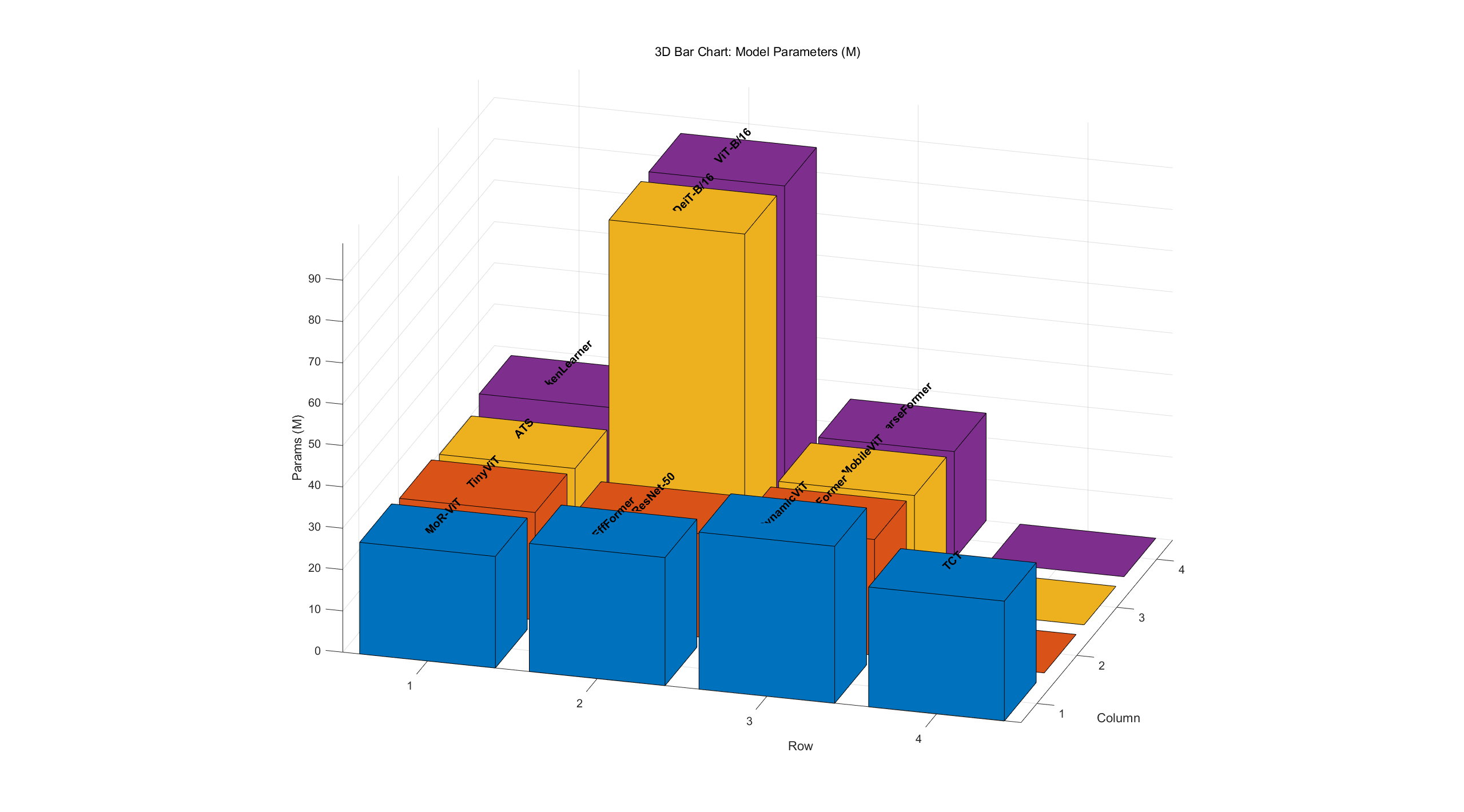}
  \caption{Comparison of model parameters among different methods (smaller is better).}
  \label{fig:MB}
\end{figure}

As shown in Fig.~\ref{fig:MB}, \textbf{IMC-Net} attains substantial parameter reduction compared with plain encoder baselines while remaining competitive with recent lightweight designs. This compactness lowers memory and storage costs and simplifies firmware updates and device-side deployment in practice.

\subsubsection{Computational Complexity (FLOPs)}
\begin{figure}[H]
  \centering
  \includegraphics[width=0.8\linewidth]{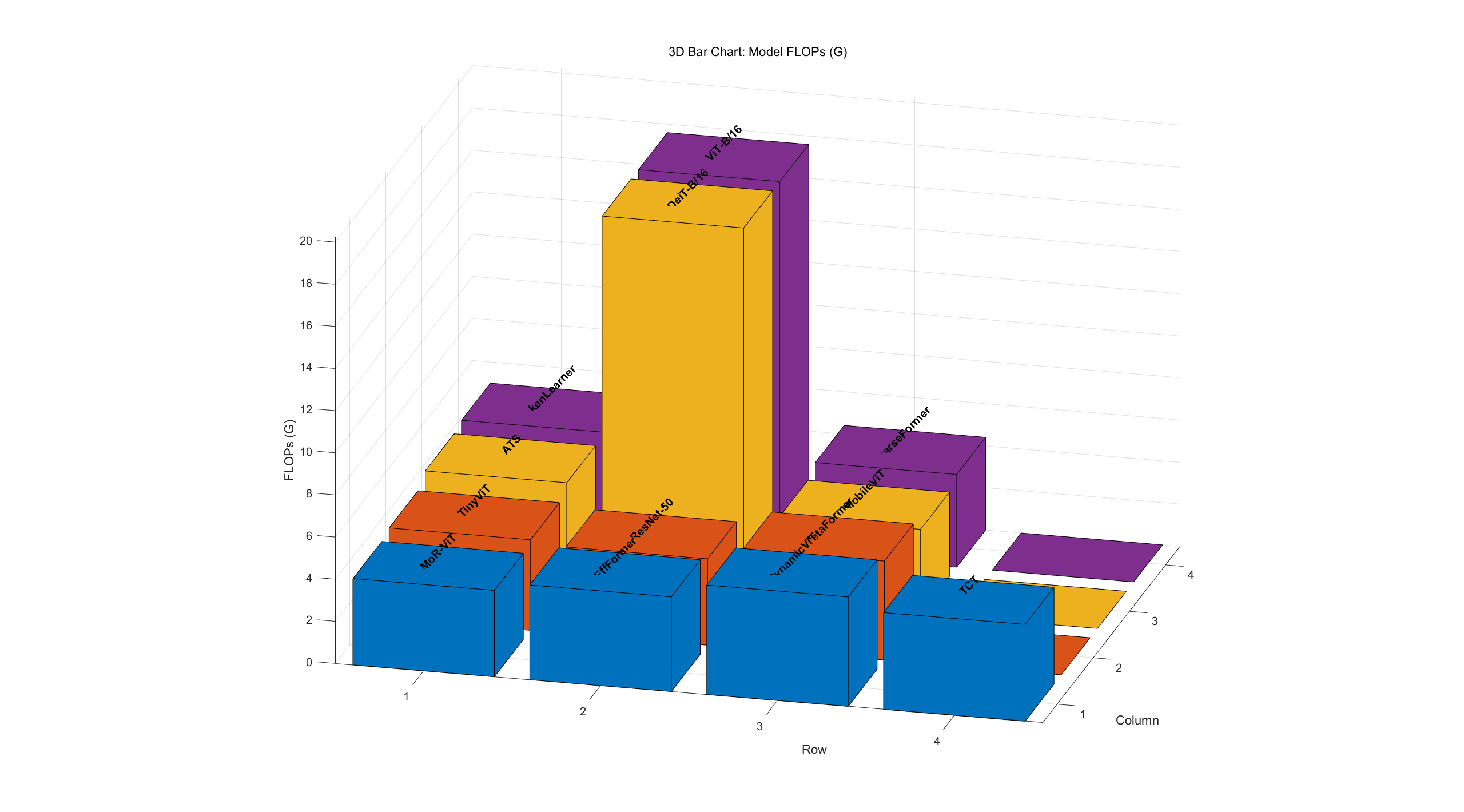}
  \caption{Comparison of FLOPs among different methods (lower is better).}
  \label{fig:flop}
\end{figure}

As illustrated in Fig.~\ref{fig:flop}, \textbf{IMC-Net} operates with consistently lower FLOPs than mainstream encoder counterparts, enabling efficient execution on resource-limited hardware. The reduced arithmetic workload also helps curb power draw and overall operational costs during large-scale deployment.

\subsubsection{Inference Throughput}
\begin{figure}[H]
  \centering
  \includegraphics[width=0.8\linewidth]{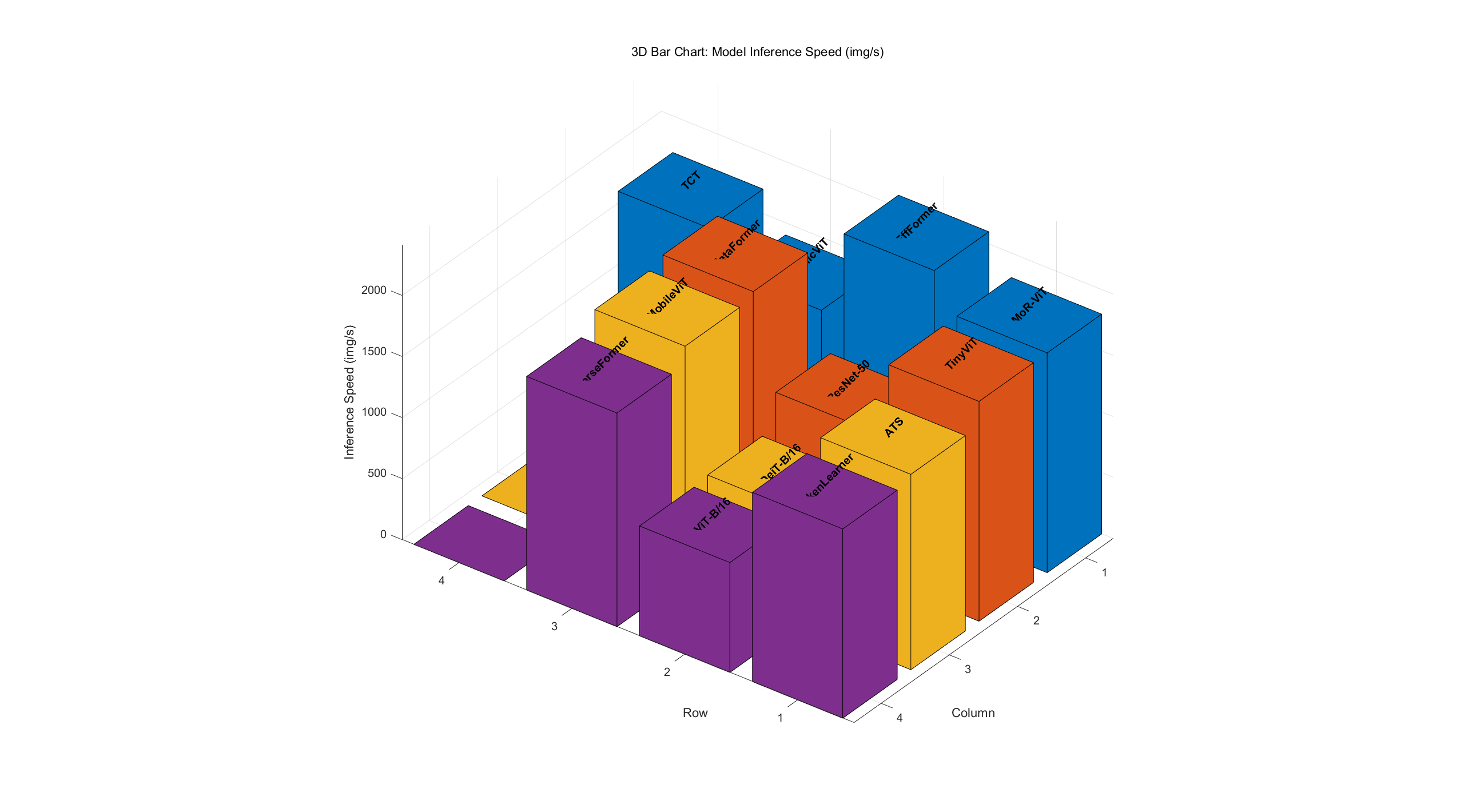}
  \caption{Comparison of inference speed among different methods (higher is better).}
  \label{fig:img}
\end{figure}

The content-conditioned \emph{multi-pass} design brings a tangible advantage in throughput (Fig.~\ref{fig:img}). By allocating extra processing only to regions estimated as complex, \textbf{IMC-Net} sustains high images-per-second without compromising recognition quality, making it suitable for scenarios that demand both rapid and reliable predictions.

\subsubsection{Generalization and Transferability}
\begin{figure}[H]
  \centering
  \includegraphics[width=0.6\linewidth]{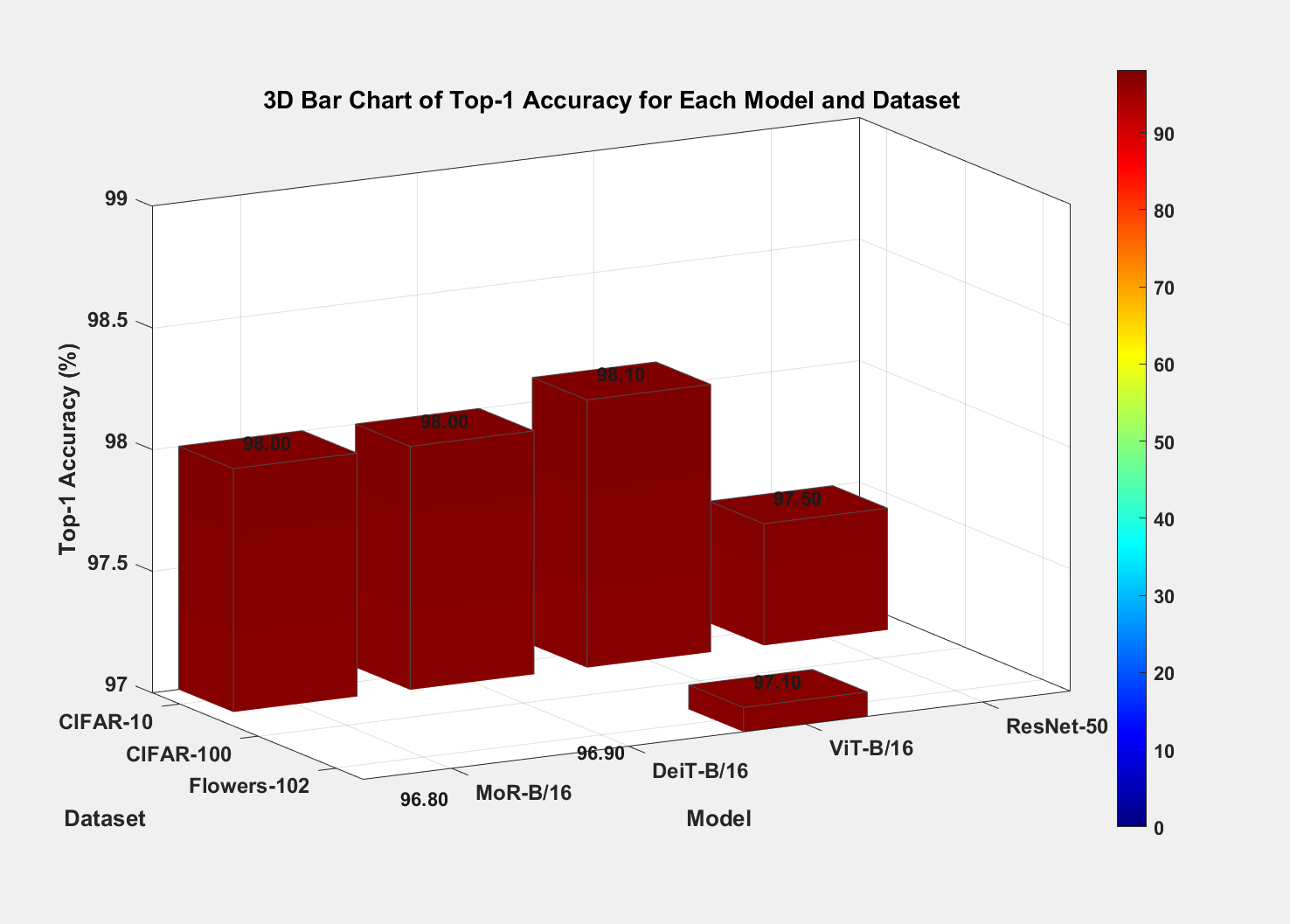}
  \caption{Top-1 accuracy on downstream benchmarks (higher is better).}
  \label{fig:acc_bar3}
\end{figure}

We further assess cross-domain behavior on CIFAR-10/100 and Flowers-102. As visualized in Fig.~\ref{fig:acc_bar3}, \textbf{IMC-Net} is competitive across coarse- and fine-grained classification tasks, matching or surpassing strong encoder baselines. The consistent performance suggests that the compact architecture and selective multi-pass control transfer well without task-specific customization.

\section{Conclusion}

This work introduced \textbf{IMC-Net}, a compact encoder that combines content-conditioned \emph{multi-pass} processing with lightweight region-wise selection and minimal module reuse. The design departs from uniform depth pipelines by assigning additional passes only where the input appears complex, while keeping the architecture small and the implementation deployment-friendly.

Across ImageNet-1K and several downstream benchmarks, IMC-Net delivers competitive recognition quality with up to 68\% fewer parameters and about $2.0\times$ higher throughput under comparable settings. These gains are obtained without external large-scale pretraining or distillation. Controlled ablations indicate that (i) region-wise selection for multi-pass control, (ii) compact core re-application, and (iii) mild score regularization jointly account for the favorable accuracy–efficiency trade-off.

Looking ahead, the same principles—selective additional passes, compact cores, and stable score shaping—can be extended to larger scales, broader visual tasks, and edge scenarios. Future directions include automated pass-budget policies, integration with architecture search, and theoretical analysis of pass allocation. We hope IMC-Net provides a practical step toward scalable, deployment-ready visual recognition systems.

\newpage
\bibliographystyle{IEEEtran} 
\bibliography{reference}      
\end{document}